\documentclass[sigconf]{acmart}

\makeatletter
\renewcommand\@authorfont{\small}
\renewcommand\@affiliationfont{\footnotesize}
\makeatother

\AtBeginDocument{%
  }

\usepackage{xurl}

\acmConference[Loco '26]{2nd International Workshop on Low Carbon Computing}{Lancaster, UK}

\begin{document}
\setcopyright{none}
\renewcommand\footnotetextcopyrightpermission[1]{}
\pagestyle{plain}
\settopmatter{printacmref=false, authorsperrow=4}
\title{Green BOA: Determining the environmental break-even point for ML-based data compression}


\author{Caterina Doglioni}
\email{caterina.doglioni@manchester.ac.uk}
\affiliation{%
  \institution{University of Manchester}
  \city{Manchester}
  \country{UK}
}

\author{Akshat Gupta}
\email{akshat.gupta-4@postgrad.manchester.ac.uk}
\affiliation{%
  \institution{University of Manchester}
  \city{Manchester}
  \country{UK}
}

\author{Thomas Elliott}
\email{thomas.elliott@manchester.ac.uk}
\affiliation{%
  \institution{University of Manchester}
  \city{Manchester}
  \country{UK}
}

\author{Hanzila Hussain}
\email{hanzilahussain234@gmail.com}
\affiliation{%
  \institution{University of Manchester}
  \city{Manchester}
  \country{UK}
}

\author{Sanjiban Sengupta}
\email{sanjiban.sengupta@cern.ch}
\affiliation{%
  \institution{CERN/University of Manchester}
  \city{Manchester}
  \country{UK}
}

\author{Zhengkai Sun}
\email{zhengkai.sun@student.manchester.ac.uk}
\affiliation{%
  \institution{University of Manchester}
  \city{Manchester}
  \country{UK}
}

\begin{abstract}
We summarise the outcome of two summer internship projects based at the University of Manchester, focused on the break-even point in terms of environmental sustainability for ML-based data compression algorithms. Using the example of a ML-based lossless compression algorithm, we compare estimates for the carbon-equivalent of the infrastructure needed for ML training and inference with the carbon-equivalent savings from reduced disk storage requirements, and discuss their break-even point. 
\end{abstract}

%\keywords{Sustainable computing, research computing, data compression, ML-based data compression, data storage}

\maketitle
\vskip-7pt
\textit{\textbf{Context.}} Big Data experiments, such as those at the Large Hadron Collider \cite{LHC} and in other astroparticle physics experiments (e.g. \cite{Scaife2020}), will be recording several Exabytes of data. 
This has a significant cost in terms of both budget \cite{CERN-LHCC-2022-005} and environmental resources for data storage (see e.g. \cite{Tannu_2023,packer2025carbon}). R\&D on data compression is ongoing, and includes machine learning (ML-) based data compression. Since this kind of compression techniques are generally more computationally intensive than standard compression algorithms such as ZSTD~\cite{collet2021zstd},and LZMA~\cite{pavlov2007lzma}, we investigate the break-even point between the $CO_2$-equivalent cost of training and executing the ML-based compression algorithm, and the embodied pllus operational $CO_2$-equivalent of disk storage that would be displaced by storing data in its compressed form. 

\vskip3pt
\noindent \textit{\textbf{Methods.}} We consider the energy use of a ML-based data compression algorithms developed at the University of Manchester \cite{Gupta_2026}, and estimate its energy usage for ML training and inference (compression/decompression round trip) on an Nvidia T4 GPU. We consider several country-specific scenarios for where these steps will be executed, to convert energy consumption into a $CO_2$-equivalent based on average energy mix. 

We then consider the carbon-equivalent cost of manufacturing and operating disk storage devices, in the shape of Hard Disk Drives (HDD) and tapes, for a 5 years lifetime. As examples, we consider Seagate Exos X18 HDDs \cite{seagate2023exosx18} and tape cartridges \cite{johns2021tape} (following the methodologies in \cite{packer2025carbon,vanderbauwhede2025lca}), hypothetically located in the UK. 

Inspired by \cite{10.1109/ASE63991.2025.00139}, we define the break-even point for this project as the \textit{size of the uncompressed dataset at which the carbon cost of training and running the compression algorithm equals the embodied carbon of the extra disk storage needed for the uncompressed data}. Further information on the methods can be found in the Appendix.

Rather than a full analysis, this is a proof-of-principle calculation with several assumptions using a limited amount of data.   
Only operational costs from CPU, GPU and RAM are considered for the ML-based data compression, as we assume that the GPU will already be available and that its use for ML-based data compression will be negligible compared to other uses within a LHC experiment. 
The \textbf{scenario} considered for the ML-based data compression comprises training on the full dataset, followed by one compression and one decompression round (on an existing temporary disk). A broader range of scenarios will be explored in future work.  
Since training dominates the energy consumption, this can be considered an upper bound in terms of energy consumption, as in practice BOA's generalisation capability outlined in \cite{Gupta_2026} allows for training only on a small dataset and then using this trained model for a much larger dataset. 

\vskip3pt
\noindent \textit{\textbf{Results and conclusions.}}

\begin{figure}[h]
    \centering
    \includegraphics[width=\linewidth]{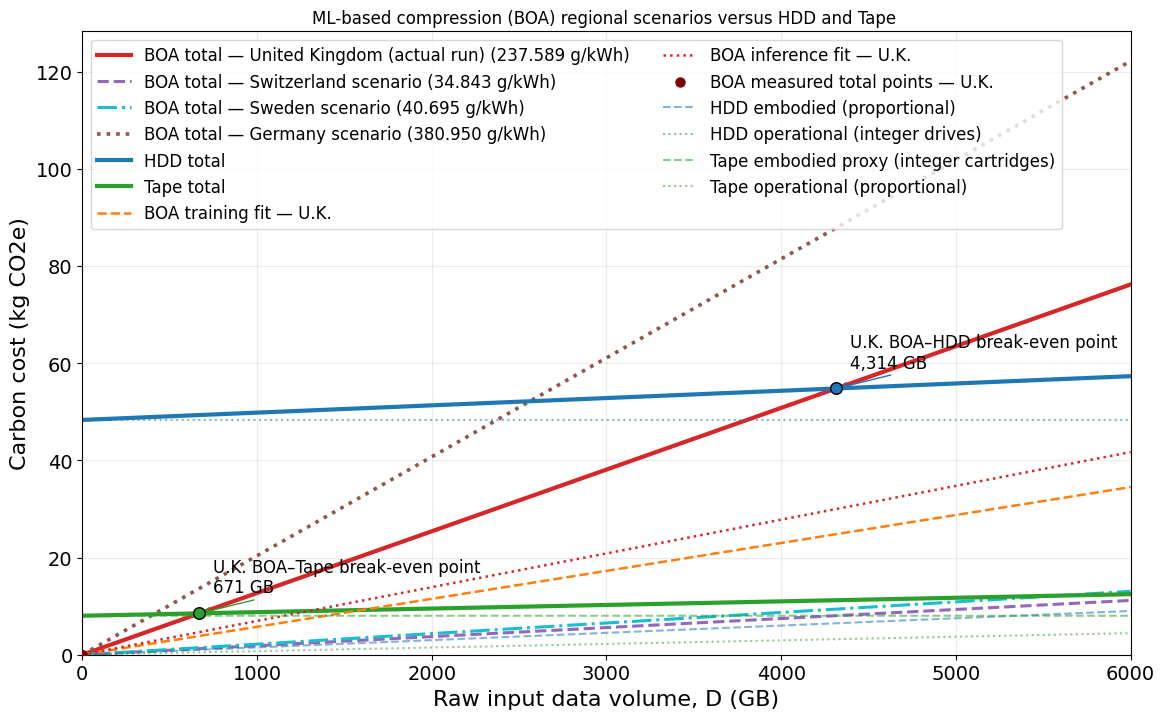}
    \caption{Results of the break-even analysis in terms of $CO_2$-equivalent for ML-based data compression and subsequent displaced HDD/tape storage.
    }
    \label{fig:boa-results}
\end{figure}

As it can be seen in Figure \ref{fig:boa-results}, the location of the break-even point is very sensitive to the carbon intensity of the country where the ML-based data compression is performed. As shown in previous literature, tape storage has a lower $CO_2$-equivalent footprint than HDDs, but at the cost of slower data access.

\vskip-7pt

\vskip-13pt

We also compared ML-based data compression to standard algorithms, taking performance (compression ratio) into account: since BOA generally achieves a better compression ratio but at lower throughput with respect to standard algorithms \cite{Gupta_2026}, it will be more $CO_2$-intensive per unit of data processed than non-ML algorithms.
While the team is working on improving the throughput, it will still be worth examining whether the compression gains of ML-based approaches justify the additional environmental cost in deployment at scale.

\clearpage

\section*{Appendix: technical configuration}

In the following, we detail the computing hardware and software configuration used to obtain these results, as well as the setup for the estimate of HDD and tape $CO_2$-equivalent. 

\subsection*{ML data compression}

The BOA ML compression algorithm was trained and executed on a Google Colab notebook with a runtime Nvidia T4 GPU. 
The model chosen was a two-layer Mamba-v1 backbone with a hidden dimension of 64, a byte vocabulary of 256 symbols, a sequence length of 10,000 bytes and a batch size of 5. 
Training was performed for eight epochs in FP32 precision, using a learning rate of \(5\times10^{-4}\)and a random seed of 42.
The training step included the eight training epochs, validation, final test evaluation and checkpoint writing. The compression step included the execution of one compression and decompression round-trip. 
The file that was compressed was the 49.92 MB "bundled CMS file" (\texttt{CMS\_DATA\_float32.bin}) from \cite{Gupta_2026} that can be found within the BOA GitHub repository \cite{gupta_2026_18481928}; each data record contains 24 \textsc{float32} features. The results from this file were scaled to the maximum dataset size shown in Figure \ref{fig:boa-results} after checking linearity of training and inference carbon costs. 

Carbon tracking was performed using CodeCarbon \cite{benoit_courty_2026_21791294}, following a tutorial in \cite{hannaTrackingEmissions}. CodeCarbon was used in one-second power-sampling intervals using \textit{machine} mode to measure the energy consumption of CPU, GPU and RAM used in the ML-based compression training and inference. The same setup was used for the compression-decompression round trip using ZSTD and LZMA. We are aware that \textit{machine} mode measures the energy consumption of the hardware stack of GPU, CPU and RAM (which may as well be shared when using a Google Colab notebook), but it is the only way to obtain a GPU energy estimate using CodeCarbon. Thermal Design Power (TDP) scaling for an Intel(R) Xeon(R) CPU @ 2.00GHz was used for the CPU power consumption corresponding to 8W, while the GPU power consumption used the \textit{nvidia-ml-py} package. 

The conversion between energy use and carbon equivalent was performed using CodeCarbon's national energy mix averaged scenarios \cite{codecarbonEnergyMix2023}.

\subsection*{Storage servers}

The carbon cost included both embodied and operational carbon \cite{packer2025carbon,vanderbauwhede2025lca}. 

For the storage of D GB over T years on a medium m, the total storage carbon cost is
\begin{equation}
C_m(D,T)
=
C_{m,\mathrm{embodied}}(D)
+
C_{m,\mathrm{operational}}(D,T).
\end{equation}

For HDD storage, a drive with usable capacity $K_{HDD}$ is the basic hardware unit. The number of powered drives required to store D GB is
\begin{equation}
N_{\mathrm{HDD}}(D)
=
\max\left(
N_{\mathrm{HDD,min}},
\left\lceil
\frac{D}{K_{\mathrm{HDD}}}
\right\rceil
\right),
\end{equation}
where we choose $N_{\mathrm{HDD,min}}=1$, meaning at least one drive is assumed to remain installed and operational. Given the limited scope and total data volume of this study, the displaced carbon (\textit{non-use}) for the HDD is allocated proportionally to the storage capacity
\begin{equation}
C_{\mathrm{HDD,non\text{-}use}}(D)
=
D\,e_{\mathrm{HDD,non\text{-}use}},
\end{equation}
where$e_{\mathrm{HDD,non\text{-}use}}$ is the capacity-allocated displaced carbon in $gCO_2e/kWh$. The operational carbon was calculated using the integer number of powered drives
\begin{equation}
C_{\mathrm{HDD,operational}}(D,T)
=
N_{\mathrm{HDD}}(D)
\frac{P_{\mathrm{HDD}}}{1000}
(24)(365.25)T
I_{\mathrm{HDD}}
\mathrm{PUE},
\end{equation}
we neglect the Power Usage Effectiveness (PUE) throughout (and set it equal to unity).
So the total HDD carbon cost is the sum of these two quantities. 
The HDD scenario used the Seagate Exos X18 18 TB drive. 
Its nominal capacity and average idle power were taken as 18 TB and 5.3 W, respectively \cite{seagate2023exosx18}. 
The displaced $kgCO_2e$ value was taken as 27 $kgCO_2e$ per drive, corresponding to a proportionality factor of 1.50 $gCO_2e/GB$. 

For Tape storage, cartridge-related non-use carbon was calculated using an integer number of LTO-8 cartridges, while operational carbon was calculated as proportional to the complete RAL Tape-service factor in \cite{packer2025carbon}. While inconsistent with the choice made for HDDs, this choice reflected existing data availability. 

Similarly to HDD, the number of cartridges is
\begin{equation}
N_{\mathrm{Tape}}(D)
=
\max\left(
1,
\left\lceil
\frac{D}{K_{\mathrm{Tape}}}
\right\rceil
\right).
\end{equation}

The total carbon cost is calculated as 
\begin{equation}
C_{\mathrm{Tape}}(D,T)
=
N_{\mathrm{Tape}}(D)
E_{\mathrm{Tape,non\text{-}use}}^{\mathrm{cartridge}}
+
\frac{D}{1000}\,
T\,
e_{\mathrm{Tape,RAL}},
\end{equation}
where $E_{\mathrm{Tape,non\text{-}use}}^{\mathrm{cartridge}}
$ is the non-use lifecycle proxy for one cartridge in $gCO_2e/cartridge$, and $e_{\mathrm{Tape,RAL}}$ is the complete RAL Tape-service operational factor in $gCO_2e/GB year$. 
The first term is cartridge-related displaced (\textit{non-use}) carbon, while the second represents operational emissions allocated to the stored data volume.
The model used an LTO-8 cartridge with a native capacity of 12 TB. 
The corresponding lifecycle data report a total value of $13.72 kgCO_2e$ per cartridge, of which 5.70 $ kgCO_2e$ is assigned to the operational phase \cite{johns2021tape,packer2025carbon}. 
\begin{equation}
E_{\mathrm{Tape,non\text{-}use}}^{\mathrm{cartridge}}
=
E_{\mathrm{Tape,total}}^{\mathrm{cartridge}}
-
E_{\mathrm{Tape,use}}^{\mathrm{cartridge}}.
\label{eq:tape-non-use-proxy}
\end{equation}
Here,
$E_{\mathrm{Tape,total}}^{\mathrm{cartridge}}$
is the reported total lifecycle carbon emission of one tape cartridge, and
$E_{\mathrm{Tape,use}}^{\mathrm{cartridge}}$
is the corresponding use-phase contribution. Using the literature values of
\(13.72~\mathrm{kgCO_2e/cartridge}\) and
\(5.70~\mathrm{kgCO_2e/cartridge}\), respectively \cite{johns2021tape,packer2025carbon}, gives
\begin{equation}
E_{\mathrm{Tape,non\text{-}use}}^{\mathrm{cartridge}}
=
13.72-5.70
=
8.02~\mathrm{kgCO_2e/cartridge}.
\label{eq:tape-non-use-value}
\end{equation}

\begin{acks}
This work was funded by EPSRC Studentship EP/W524347/1 (Project 2932638) via MADSIM CDT, the University of Manchester Dame Kathleen Ollerenshaw Fellowship, and is part of a project that has received funding from the European Research Council under the European Union’s Horizon 2020 research and innovation program (grant agreement 101002463). Summer student funding was also provided by the N8CIR summer internship scheme.  
\end{acks}

\bibliographystyle{ACM-Reference-Format}
\bibliography{references}

\end{document}